%% file: main.tex
\documentclass[conference]{IEEEtran}
\IEEEoverridecommandlockouts
\usepackage[utf8]{inputenc}
\usepackage{xcolor}
\usepackage{hyperref} 
\usepackage{url}
\usepackage{verbatim}
\usepackage{cite}
\usepackage{longtable}
\usepackage{pifont}   
\usepackage{multirow} 
\usepackage{amsmath,amssymb,amsfonts}
\usepackage{algorithmic}
\usepackage{graphicx}
\usepackage{textcomp}
\usepackage{xcolor}
\def\BibTeX{{\rm B\kern-.05em{\sc i\kern-.025em b}\kern-.08em
    T\kern-.1667em\lower.7ex\hbox{E}\kern-.125emX}}

\begin{document}

\title{
Right-sizing Recommendations (RSR): Cloud Workload Conformal Prediction for Virtual Machines in Data Center Operations

\thanks{© 2025 IEEE. Personal use of this material is permitted. Permission from IEEE must be obtained for all other uses, including reprinting/republishing this material for advertising or promotional purposes, collecting new collected works for resale or redistribution to servers or lists, or reuse of any copyrighted component of this work in other works.}
}

\author{
    Mehryar Majd\textsuperscript{1,2,*},
    Feng Cheng\textsuperscript{1,2},
    Ali Pahlevan\textsuperscript{2},
\\
\IEEEauthorblockN{
\textsuperscript{1}Hasso Plattner Institute (HPI), University of Potsdam, Potsdam, Germany\\
\textsuperscript{2}SAP SE, Walldorf, Germany\\
\textit{*Corresponding author: mehryar.majd@\{hpi.de, sap.com\}}
}
}
\maketitle              

\input{Chapters/1_abstract}
\input{Chapters/2_0_introduction}

\input{Chapters/2_1_BG_motivations}

\input{Chapters/2_2_Proposed_Solution_Overview}
\input{Chapters/2_3_Organization_of_the_paper}
\input{Chapters/3_0_Related_Works} 
\input{Chapters/4_0_Background_Knowledge} 
\input{Chapters/4_1_benchmarks}
\input{Chapters/4_2_CP-and-PI} 
\input{Chapters/4_3_Datasets}
\input{Chapters/5_0_Methodology}
\input{Chapters/5_1_Problem_Statement}
\input{Chapters/8_discussion_Results}
\input{Chapters/9_future-work-and-conclusion}

{\footnotesize
\bibliographystyle{unsrt} 
\bibliography{References.bib}
}

\end{document}

%% file: Chapters/1_abstract.tex
\begin{abstract}
Managing cloud infrastructure efficiently, especially in environments of large cloud providers or hyperscalers, requires optimizing the use of physical resources to minimize costs and maximize performance. Selecting the right virtual machine (VM) sizes is crucial to achieving cost efficiency in these dynamic environments. However, traditional VM allocation and scheduling approaches often fail to account for the fluctuating, unpredictable nature of VM utilization, leading to inefficiencies such as over- or under-provisioning of resources. High-quality interval prediction helps accurately capture uncertainty in cloud resource demand and supports cloud operators in efficient instance provisioning. As an effective and reliable framework for constructing prediction intervals (PIs), conformal prediction (CP) is used for mid- and long-term forecasting tasks in cloud computing environments. This study proposes a new data-driven PI construction approach using bootstrapping conformal prediction for modern/dynamic/data-driven Right-sizing Recommendations (RSR) to enhance provisioning for diverse (web-/) application workloads on hyperscalers. By learning workload utilization patterns, identifying correlations across multiple time series, and predicting medium to long-term utilization trends, this research seeks to enhance the efficiency of cloud/data center operations using an AI/ML-based provisioning pipeline, ensuring that VM resources are allocated cost-effectively according to VM workload forecast results to meet the fluctuating demands of cloud workloads. Our study demonstrates that AI-driven models, powered by machine learning (ML) regressions using backtesting, achieve promising forecasting results for cloud resource utilization. Additionally, we ranked the selected models to introduce the global top-tier models over long-life VM candidates. This approach enhances right-sizing recommendations for dynamic cloud environments.
\end{abstract}


%% file: Chapters/2_0_introduction.tex
\section{INTRODUCTION \& BACKGROUND} \label{sec:num1}
Right-sizing virtual machines (VMs) is a persistent challenge for hyperscalers in cloud environments, given the dynamic, inconsistent, and frequently unpredictable nature of their workloads. Unlike smaller-scale deployments, hyperscalers must cope with sizeable, unpredictable fluctuations in demand across thousands of tenants and applications. Even minor inefficiencies in resource allocation can result in substantial costs, increased energy consumption, and performance issues. Traditional right-sizing methods, typically based on static thresholds or simplistic utilization models, struggle to capture dynamic workload profiles, leading to either over-provisioning, which wastes resources, or under-provisioning, which threatens to compromise service-level agreements (SLAs).

To address this, predictive analytics is essential for accurately forecasting medium- to long-term workload patterns and generating actionable right-sizing recommendations. It is important to construct accurate prediction intervals, as this allows cloud operators to account for the inherent fluctuations and unpredictability of cloud workloads. Conformal Prediction (CP) provides a principled framework for constructing such intervals, yielding reliable forecasts that facilitate more precise provisioning, resource scaling, and cost optimization.
\begin{figure}[t]
\centering
\includegraphics[width=\columnwidth]{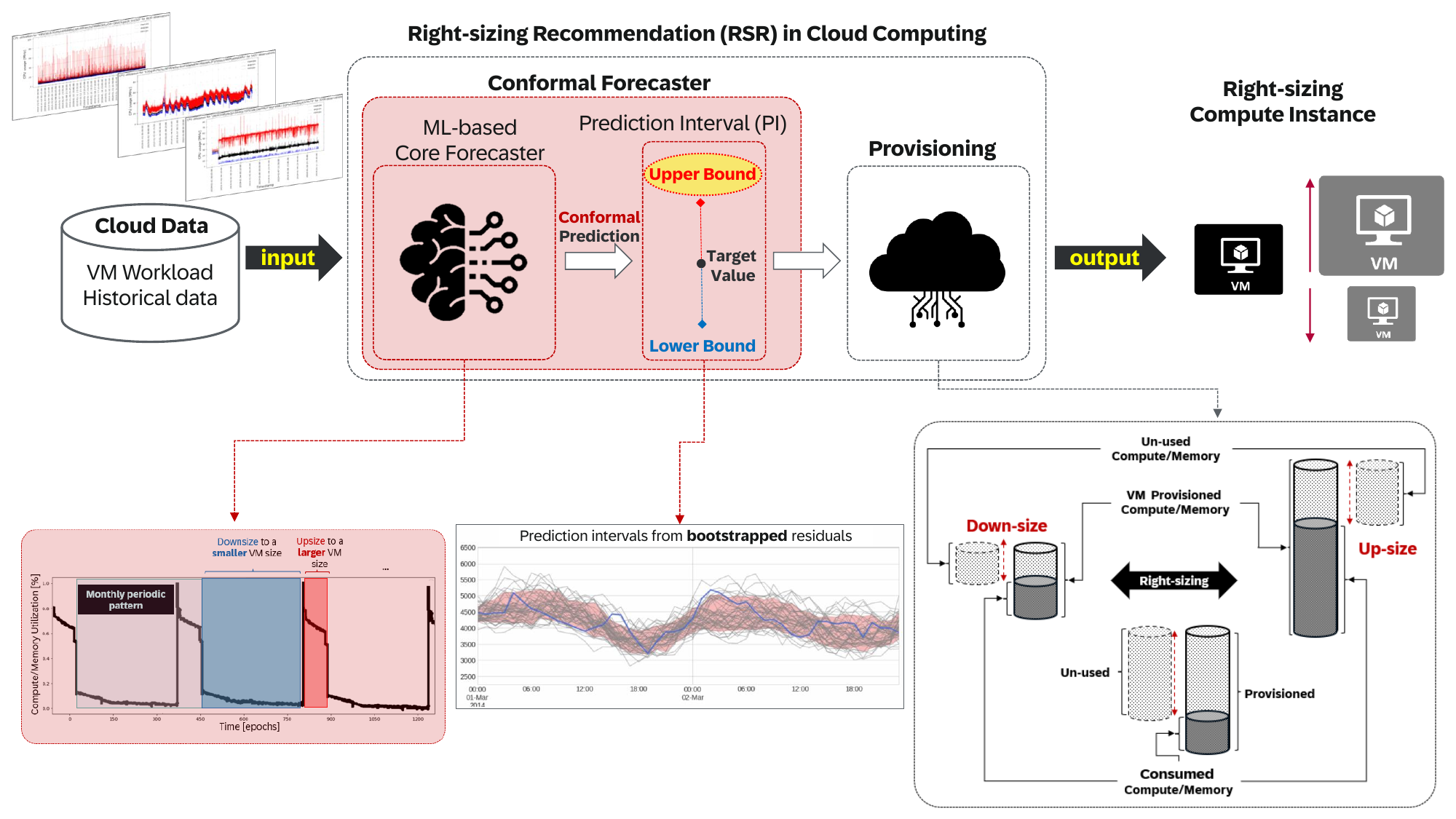} 
\caption{Overview of the proposed Right-sizing Recommendation (RSR) framework in cloud computing.
Historical VM workload data is fed into a Conformal Forecaster that generates Prediction Intervals (PIs) around the predicted target using Conformal Prediction (CP). These intervals inform the provisioning stage, which selects the optimal compute instance size (e.g., vXPU) to match future demand. The red-shaded section highlights that the core contribution of this work lies in mid-/large-scale workload prediction using CP to construct PIs that enhance cloud workload management (e.g., vCPU, vMem, vDisk) and resource optimization for the RSR use case, complemented by insights from ranking ML-based forecasting models across diverse VM workloads to inform best-practice provisioning decisions. 
}
\label{fig:1}
\end{figure}
A notable example is Resource Central (RC), a system proposed by Cortez et. al. \cite{cortez2017resource} that leverages detailed characterization of Microsoft Azure’s production VM workloads to improve resource management. Their study showed that VM behaviors often remain consistent over time, enabling historical telemetry to support accurate workload prediction. RC collects and learns from high-resolution CPU and memory usage traces offline, then provides online predictions to various resource managers to support decisions such as intelligent oversubscription. With the evolution of (Gen-)AI and modern forecasting techniques, Instance Behavior Analytics (IBA) has become increasingly effective, further motivating our focus on Right-sizing Recommendations (RSR), where predictive analytics can drive more accurate, uncertainty-aware VM provisioning in hyperscalers' environments.

%% file: Chapters/2_1_BG_motivations.tex
\subsection{Motivation and Contributions}\label{sec:num3}
Accurately estimating the upper bound (UB) of a VM’s future resource consumption is essential for right‑sizing and selecting the optimal VM configuration to meet workload needs without over‑ or under‑provisioning. When downsizing to a smaller VM, future demands must remain within the reduced memory and CPU capacities to avoid performance degradation and SLA violations.

We adopt a constraint-driven approach for the provisioning stage: after downsizing, peak memory usage should remain below 60\% of capacity, while the 99th percentile of CPU utilization should not exceed 70\%, accounting for CPU overshoots. While a naive UB estimate can be obtained from the historical maximum, this is often overly conservative, especially for workloads with periodic patterns. Instead, uncertainty-aware forecasting can provide more accurate and robust UB estimates, enabling cost-efficient, reliable right-sizing in large-scale cloud environments.

The main contributions of this paper are summarised below:
\begin{itemize}\setlength\itemsep{0.5em}
    \item We propose an AI/ML-based forecasting pipeline using CP for cloud workload prediction, extendable to the Right-Sizing Recommendation (RSR) use case.
    \item We experimentally compare the predictive accuracy of several time-efficient regression models for mid-term and long-term forecasting.
    \item We present best practices for workload prediction, leveraging backtesting (BT)-based cross-validation tailored for time-series data.
    \item We rank forecasting models across research questions to identify global top-tier models for long-lived VM workload profiles.
\end{itemize}

%% file: Chapters/2_2_Proposed_Solution_Overview.tex
\subsection{Proposed Pipeline for RSR use case}
In this research, historical CPU workload data were modeled as univariate time-series for analysis and forecasting. Different prognostic models were applied to predict high-resolution CPU observations within the instances. Using time-series data, regressor models within the Forecaster sub-pipeline are deployed to develop high-resolution predictive models for CPU workload in VMs, leveraging both current and historical CPU utilization data. 

At the end of the RSR pipeline, the Provisioner sub-pipeline is used with a suitable strategy to classify instance candidates as compute-intensive or not for right-sizing. Apart from provisioning tools that were once used with different statistical-based thresholds for right-sizing, the primary focus of this study is to compare and identify the best ML regression models that can efficiently fit the nature of the VM workload. In our case, we investigated CPU data from public datasets, uniformly organized by Microsoft Azure VM family as it was recorded. In the motivation sub-section, we explained some practical strategies for high-resolution CPU/memory workload. Here, we focused solely on the VMs' CPU utilization data. 

In this study, potential ML-based prognostic models listed in Table~\ref{tab:regressors}, empowered by the Prediction Interval (PI) technique, were trained to predict the CPU workload PI and ultimately monitor the upper bound utilization on time-series data. The best predictive model was selected based on error analysis. The predictive models' performance was assessed through error analysis, and various prognostic metrics were calculated and reported. Additionally, a visualized evaluation method was proposed and implemented, along with an independent error analysis to validate the results of target-point forecasting metrics using \texttt{scikit-learn} Python packages, as well as PI forecasting metrics.  

\begin{figure}[h]
\centering
\includegraphics[width=\columnwidth]{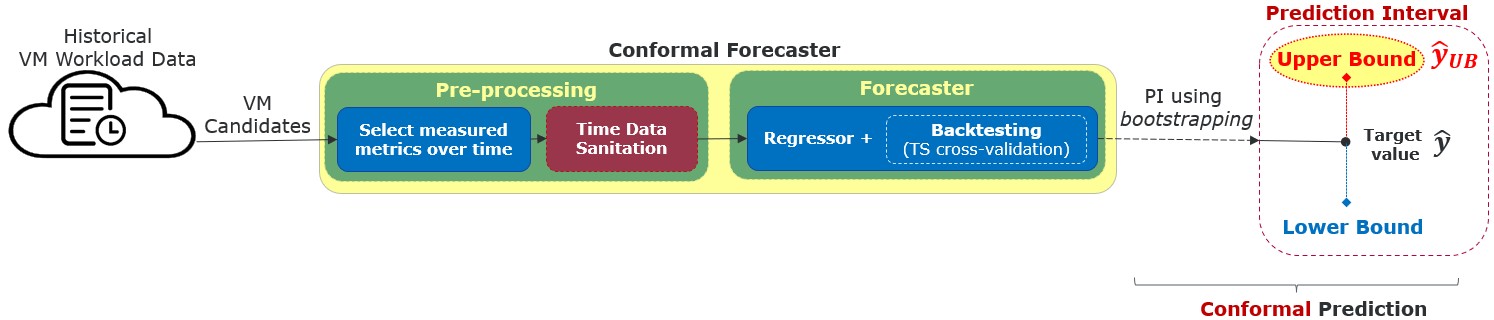}
\caption{
The excerpt of the proposed pipeline architecture, empowered by PI and ultimately monitors the upper bound (UB) utilization on time-series data for the RSR use case, enables probabilistic forecasts of the nature of VM workload for a mid/long-term horizon that could be deployed for provisioning decisions. 
}\label{fig:2}
\end{figure}
 

%% file: Chapters/2_3_Organization_of_the_paper.tex
\subsection{Organization of the Paper}
The remainder of this paper is organized as follows:  
Section~\ref{sec:num1} introduces the problem context, background, and motivations. Section~\ref{sec:num2} reviews existing literature on VM workload forecasting, RSR approaches, and conformal prediction techniques. Section~\ref{sec:num3} outlines the theoretical foundations, benchmark methods, and principles of CP and prediction intervals, as well as public datasets and their characteristics. Section~\ref{sec:num5} provides a detailed description of the problem statement and proposed methodology. Section~\ref{sec:num8} presents and discusses the experimental results. Finally, Section~\ref{sec:num9} concludes the paper and outlines directions for future research.

%% file: Chapters/3_0_Related_Works.tex
\section{RELATED WORKS}\label{sec:num2}
Our literature review, compiled from a recent survey of workload prediction studies in cloud data centers, reveals that the majority of research has focused on target (point) prediction approaches, with a noticeable peak in publication activity in 2021. Although these methods dominate the literature, conformal prediction techniques, which provide interval-based predictions with statistical guarantees, remain underexplored. Their adoption appears highly dependent on specific use cases, and their representation in the surveyed body of work is limited to a small fraction of studies. This suggests that, despite growing interest in reliable, uncertainty-aware prediction, conformal prediction remains an under-investigated area in the context of large-scale workload prediction in cloud environments. \cite{yekta2023review, bliedy2025resource}

A very recent survey explored the importance of an application-oriented perspective for cloud workload prediction in proactive resource management, highlighting its role in supporting performance assurance, cost reduction, and energy optimization. The authors systematically categorize prediction approaches based on workload variability and heterogeneity, and their work reviews closely related studies and frameworks of proactive application AIOps (Artificial Intelligence for IT Operations). Furthermore, they outlined several useful workarounds and areas for future research to address the challenges of large-scale workload prediction. \cite{feng2024application}

Another recent study introduced \textit{Pitot} frames workload runtime prediction as an interference-aware matrix completion problem, introducing a log-residual training objective, interference-aware matrix factorization, and conformalized quantile regression for uncertainty quantification, achieving more accurate and tightly bounded predictions than existing methods.\cite{huang2025interference}

\cite{rossi2025forecasting} proposed a framework for evaluating deep learning-based probabilistic forecasting models, assessing univariate and bivariate HBNN and LSTMD architectures on CPU, GPU, and memory usage prediction using Google Cloud and Alibaba traces. Unlike prior work that relies on pointwise accuracy metrics, they emphasize uncertainty-aware predictions, demonstrating benefits for cloud service performance. They also investigate transfer learning to different-distribution domains, finding performance degradation across cloud providers and limited benefits from fine-tuning. However, large and diverse source domains yield more robust generalization. Confidence levels can positively influence key cloud service performance metrics.

\begin{figure}[ht]
\centering
\includegraphics[width=\columnwidth]{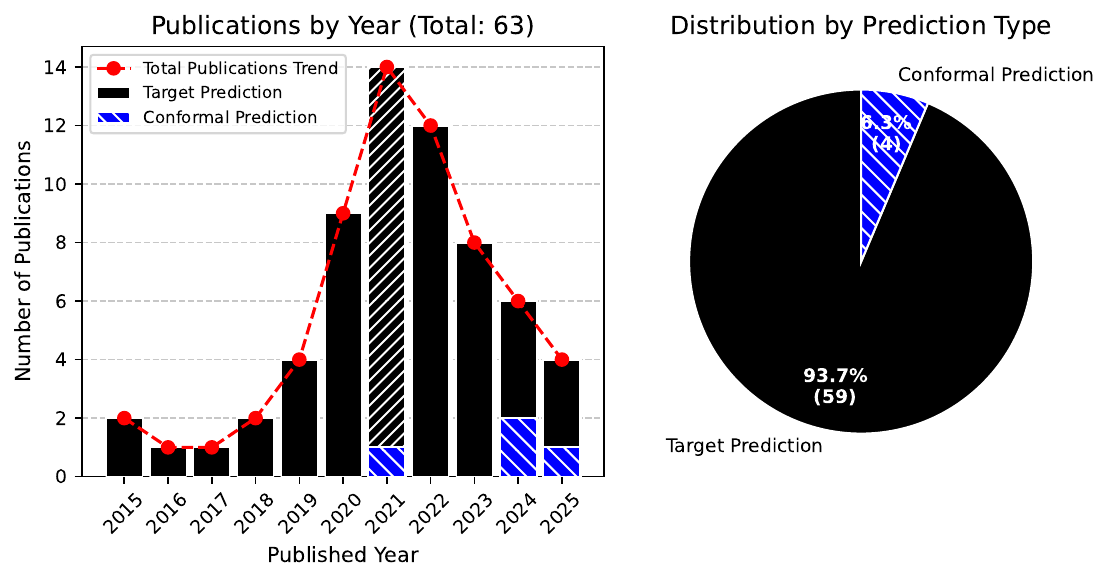}
\caption{
Annual publication trends (left) and proportions by prediction type (right) based on recent surveyed workload prediction studies on cloud workload forecasting in cloud datacentres, showing the dominance of target prediction (93.7\%) over conformal prediction (6.3\%).
}\label{fig:3}
\end{figure}

Previous research has explored resource forecasting in private clouds using classical, ensemble, and deep learning methods for predicting CPU usage. Studies indicate that the distribution of prediction types, as shown in the pie chart in Figure~\ref{fig:3}, demonstrates better performance than other models and surpasses naïve baseline approaches. Using multivariate inputs, such as CPU and memory, yields slight improvements in accuracy while reducing the risk of underprediction. Additionally, uncertainty quantification through Quantile Regression and Conformalized Quantile Regression provides well-calibrated, tighter prediction intervals compared to the overly conservative standard Conformal Prediction.\cite{PerssonSuorra2025}
Conformal Prediction (CP) remains underexplored, particularly for VM workload forecasting at mid- to large-scale, leaving a gap for right-sizing in cloud environments, as shown in the statistical plots in Figure~\ref{fig:3}.

An AI/ML-driven approach can potentially help estimate the long-term profiles of VM candidates, enabling us to enhance the right-sizing mechanisms previously based on statistics, thresholds, or rules. While some studies have employed predictive models based on artificial neural networks (ANNs), recent advances in deep neural network (DNN) architectures have yielded greater predictive accuracy. These models can effectively learn from time-series data on resource utilization and extract potential CPU consumption patterns for applications/services running on a given server or host. Developing such complex models enhances instance profiling and improves workload trace analysis for large-scale cloud platforms, such as Microsoft Azure. The recent GenAI-based approaches using LLM-based time models enable us to forecast low-frequency time data in real time (due to limitations in the input prompt's length), which is highly interesting at an enterprise scale, as it facilitates easy-to-solve tasks by generating a prompt as a query and returning the desired feedback.
On the other hand, our in-practice investigation uncovered a disconnection between provisioning tools and the VM stats collected thereafter; it is not a closed feedback system. Hence, there is no sophisticated intelligence built into the provisioning tool. This is a research gap in this domain that warrants further investigation in future work. Therefore, cloud computing resource providers or cloud service providers (CSPs) require Instance Behavior Analytics (IBA) approaches to profile workload, which leads to achieving smarter Right-Sizing Recommendations (RSRs), which could be used as Next-Gen provisioning tools for VMs, particularly in hyperscalers for RSR use cases shown in Figure~\ref{fig:2}.

%% file: Chapters/4_0_Background_Knowledge.tex
\section{Background Knowledge}\label{sec:num3}

\subsection{Microsoft Azure Virtual Machine (VM) Workload}
The Compute/Memory workload data was collected using a data collector that monitored and managed time-series data systems and was subsequently used in the RSR system. In general, workload measurements used in cloud environments to manage resource provisioning and task scheduling typically involve key performance indicators (KPIs), such as virtual G/CPU utilization (v-G/CPU) and virtual Memory usage (v-Mem), etc. In addition to these metrics, cloud systems often track other workload characteristics for more refined scheduling and provisioning. The workload is recorded by the data collector every 5 minutes within the instance. The data collector enables us to monitor VM workload changes by continuously observing changes in the application(s) running on servers or application servers, which, in turn, lead to changes in vCPU workload. Given the VM or instance size, the consumed $VM_{compute}$ and $VM_{memory}$ can be estimated for vCPU and vMem, respectively (often normalized). Equation~\ref{eq:vm_consumption} shows the estimation of consumed compute/memory utilization for VM workload measurement metrics $VM(c/m)_{\text{consumed}}$:
\begin{equation}
VM(c/m)_{\text{consumed}} = VM(c/m)_{\text{provisioned}} - VM(c/m)_{\text{unused}}
\label{eq:vm_consumption}
\end{equation}
There are various methods for estimating prediction intervals \cite{hyndman2018forecasting}. Previous studies have shown that PI can be accurately estimated using ML methods \cite{shrestha2006machine}.

%% file: Chapters/4_1_benchmarks.tex
\subsection{ML-based Regression Models}
The forecasting core within the proposed pipeline flexibly deploys a diverse set of time-friendly ML regressors, encompassing linear, non-linear, neural network, ensemble, and gradient-boosting techniques to accommodate varying temporal patterns and feature dependencies. \texttt{LinearRegression} \cite{linearregression2024} provides a baseline with interpretable linear modeling, while Support Vector Regression (\texttt{SVR}) \cite{svr2024} captures non-linear relationships. Multi-Layer Perceptron Regressor (\texttt{MLPRegressor}) \cite{mlpregressor2024} introduces the capacity for non-linear feature interaction learning through neural networks. Ensemble-based methods, such as Random Forest (\texttt{RandomForestRegressor}) \cite{rf2024}, exploit subsampling and decision-tree diversity to achieve robust predictions. Advanced gradient-boosting implementations, including Extreme Gradient Boosting (\texttt{XGBoost}) \cite{xgboost2024}, Categorical Boosting (\texttt{CatBoostRegressor}) \cite{catboost2024}, and Light Gradient Boosting Machine (\texttt{LightGBM}) \cite{lightgbm2024}, offer high predictive accuracy and efficiency, each optimizing gradient-based learning with distinct strategies that range from decision-tree enhancements to one-sided sampling. This heterogeneous set of regressors ensures the forecasting framework can adapt to a broad spectrum of workload dynamics and uncertainty profiles.
\begin{table}[h]
\centering
\caption{List of regressors and their ML techniques used in \texttt{ForecasterAutoreg} class.}
\label{tab:regressors}
\setlength{\tabcolsep}{4pt} 
\renewcommand{\arraystretch}{1.1} 
\begin{tabular}{c l l}
\hline
\# & \textbf{Regressor} & \textbf{ML Framework Technique} \\
\hline
1 & LinearRegression & Linear \\
2 & SVR & Non-linear \\
3 & MLPRegressor & Neural Nets (shallow) \\
4 & RandomForestRegressor & Embedded tree-based sub-sampling \\
5 & XGBoost & Gradient Boosting \\
6 & CatBoostRegressor & Gradient Boosting \\
7 & LightGBM & Gradient-based One-Side Sampling \\
\hline
\end{tabular}

\vspace{2pt}
\raggedright\footnotesize
All model configurations \& experimental setups are available on our GitHub:
\href{https://github.com/clevilll/MS-Azure-VM-WL-CP-Forecast-Characterization}{repository}

\end{table}

\subsection{Forecasting Approaches: Multi-Step Time Series Forecasting}\label{sec:num4.1}
We implemented the forecasting pipeline using the \texttt{scikit-learn} library and integrated seven machine learning regressors within the \texttt{Forecaster} component, tailored to the experimental scenarios defined by our research questions. To enable multi-step or sequence forecasting, the input time-series data were transformed into a supervised learning format using a series-to-supervised (STS) approach \cite{brownlee2017introduction}. This transformation was handled during the pre-processing stage via the Time Transformation sub-component.

\subsubsection{Backtesting (BT)}
Reliable backtesting (BT) methods are critical, as models evaluated without robust BT often fail to generalize to out-of-sample data \cite{skforecast_backtesting2024, joubert2024three}. To investigate the influence of BT on forecasting performance, we employed the \texttt{skforecast} package, which offers built-in support for multi-step forecasting and backtesting. In practice, time series forecasting typically aims to predict multiple future values and not just the next step $(t+1)$, but an entire horizon $(t+1, \dots, t+n)$ or a specific distant step $(t+n)$. Various forecasting strategies have been developed to support such objectives \cite{skforecast}.

To evaluate the effectiveness of conformal prediction methods in our cloud forecasting experiments, we examined four different backtesting (BT) strategies, although additional strategies exist. Each approach offers a different balance between computational cost and model adaptability over time. The evaluated strategies are summarized below:

\begin{itemize}\setlength\itemsep{0.5em}
    \item \textbf{Backtesting without refit}: The model is trained once and reused for all forecast windows, reducing computational load but limiting adaptability to temporal shifts.
    
    \item \textbf{Backtesting with refit and increasing training size (fixed origin)}: The model is re-trained at each step using all available past data, enabling adaptation to evolving trends while maintaining a fixed training start point.
    
    \item \textbf{Backtesting with refit and fixed training size (rolling origin)}: The training window moves forward in time with a fixed size, offering a balance between recency and model stability.
    
    \item \textbf{Backtesting with intermittent refit}: The model is periodically re-trained (e.g., every $k$ steps), reducing computation compared to full refitting while still addressing non-stationarity.
    
\end{itemize}

\subsubsection{Recursive Multi-Step Forecasting (RMSF)}\label{sec:num5.1}

In our experiments, we adopted the Recursive Multi-Step Forecasting (RMSF) strategy rather than the Direct Multi-Step Forecasting (DMSF) approach. This choice was motivated by the dependency structure inherent in the time series data to predict the value at time step $t_n$, the value at $t_{n-1}$ is required, which is itself unknown at prediction time. Consequently, a recursive process is employed, in which each forecasted value is used as input to predict subsequent steps. This iterative mechanism, known as recursive forecasting, was a central component of our evaluation. A comparative illustration of RMSF and DMSF strategies is provided in the following figure.

\begin{figure}[h]
\centering
\includegraphics[width=\columnwidth]{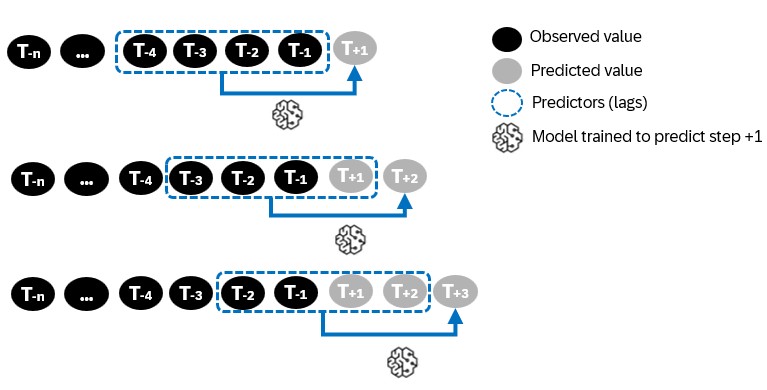}
\caption{
Diagram of recursive multi-step forecasting (RMSF). 
}\label{fig:4}
\end{figure}

The Direct forecasting strategy was not considered in our approach due to its primary drawback: the need to train a separate model for each forecast horizon, which can be computationally expensive. In contrast, the Recursive strategy requires training a single model, where each 1-step forecast is fed back into the model as a lagged input to predict subsequent steps. While more efficient, this method is prone to error accumulation over longer horizons due to the propagation of forecast errors.

To mitigate the limitations of both strategies, we employed the \textit{DirRec (Direct-Recursive)} strategy within our recursive multi-step forecasting (RMSF) framework, using a backtesting procedure based on cross-validation (CV). \cite{bergmeir2012use} The DirRec approach combines the strengths of both methods by training a separate model for each forecast step, while also incorporating previously predicted values as additional lagged features. This allows each model to maintain an updated context, capturing serial dependencies more effectively than the Direct strategy alone. However, similar to the Recursive strategy, \textit{DirRec} may still suffer from error propagation across steps.

%% file: Chapters/4_2_CP-and-PI.tex
\subsection{Conformal Prediction (CP) and Prediction Interval (PI)}\label{sec:num5.4}
A prediction interval is a quantification of the uncertainty on a prediction and is used in many domains, such as improving recommendation systems in online retail \cite{dey2023conformal}, renewable energy, and wind energy management for wind turbine power prediction \cite{gijon2025integrating}, wind speed prediction \cite{ji2008wind}, and clinical medical sciences \cite{vazquez2022conformal}, etc. It provides a probabilistic upper and lower bound on the estimate of an outcome variable. “A prediction interval for a single future observation is an interval that will, with a specified degree of confidence, contain a future randomly selected observation from a distribution.” \cite{meeker2017statistical}

Prediction intervals are most used when making predictions or forecasts with a regression model, where a quantity is being predicted. In general, Conformal Prediction (CP) could be estimated for regressions \cite{bao2025review} via: 

\begin{itemize}\setlength\itemsep{0.5em}
    \item \textbf{Conformalized Mean Regression (CMR):} (start from a point prediction) 
\begin{equation}
\hat{f}(x_i) 
\xrightarrow{\text{CP}} 
C(x_i) = \left[ \hat{f}(x_i) - q_{1-\alpha},\; \hat{f}(x_i) + q_{1-\alpha} \right]
\end{equation}
where $\hat{f}(x_i)$ is a point prediction and $q_{1-\alpha}$ is the $(1-\alpha)$-quantile of calibration residuals.
    \item \textbf{Conformalized Quantile Regression (CQR):} (start from (non-conformal) intervals produced by quantile regression) 

\begingroup
\small
\begin{equation}
\left[ \hat{f}_{\mathrm{L}}(x_i),\; \hat{f}_{\mathrm{U}}(x_i) \right] 
\xrightarrow{\text{CP}} 
C(x_i) = \left[ \hat{f}_{\mathrm{L}}(x_i) - q_{\alpha},\; \hat{f}_{\mathrm{U}}(x_i) + q_{\alpha} \right]
\end{equation}
\endgroup

where $\hat{f}_{\mathrm{L}}(x_i)$ and $\hat{f}_{\mathrm{U}}(x_i)$ are lower/upper quantile predictions, and $q_{\alpha}$ is the calibration adjustment ensuring coverage $1-\alpha$.

\end{itemize}
The Conformal Prediction (CP) transforms point predictions into statistically valid prediction intervals. We used bootstrapping to create prediction intervals with the \textit{skforecast} package, which requires only that the residuals (errors) be uncorrelated, whereas most PI methods require that the model residuals be normally distributed. Using bootstrapping process results, prediction intervals can be computed by calculating the $\alpha/2$ and $1-\alpha/2$ percentiles at each forecasting horizon.

\begin{figure}[h]
\centering
\includegraphics[width=\columnwidth]{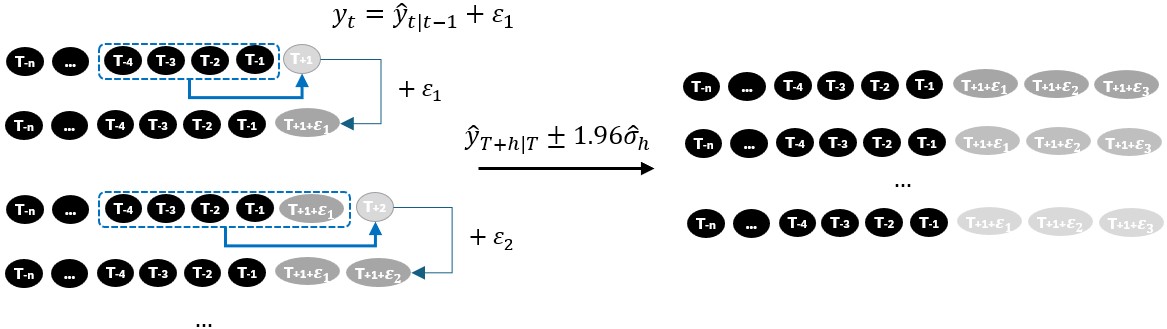}
\caption{
Diagram of the bootstrapping prediction process. 
}\label{fig:5}
\end{figure}

The bootstrapped residuals approach estimates forecast uncertainty by resampling historical prediction errors, thereby producing intervals rather than single-point forecasts. The one-step-ahead forecast error is defined as
\begin{equation}
\varepsilon_t = y_t - \hat{y}_{t \mid t-1}
\end{equation}
Assuming future errors resemble past residuals, synthetic forecasts are obtained by adding resampled errors to model predictions. Repeating this process yields a forecast distribution from which prediction intervals are computed via empirical $\alpha/2$ and $1-\alpha/2$ quantiles or by fitting a parametric distribution. This method requires only one trained model, but can be computationally intensive when many bootstrap replications are needed.

%% file: Chapters/4_3_Datasets.tex
\subsection{Datasets}
In alignment with realistic data center observations, we used publicly available Microsoft Azure Virtual Machine (VM) workload datasets curated by other researchers for workload characterization. The traces are sanitized subsets of first‑party VM workloads collected from one of Azure’s geographical regions. These datasets are particularly suitable for our mid- to large-scale forecasting objectives due to their long-term time span and the availability of workload profiles categorized by usage patterns, a feature lacking in many comparable datasets. Furthermore, to make our workaround fully reproducible, we selected these openly accessible datasets. Specifically, we used workload candidates from \texttt{AzurePublicDatasetV1} (2017) and \texttt{AzurePublicDatasetV2} (2019) and summarized them in Table~\ref{tab:2}.  The key characteristics of these datasets are documented in the associated GitHub repository\footnote{\url{https://github.com/Azure/AzurePublicDataset}}.

\begin{table}[htbp]
\centering
\caption{Public MS Azure VM workload datasets and their \textcolor{blue}{long-lifetime} VM candidates ($VM_{length} \geq 29$ consecutive days).}
\label{tab:azure_datasets}
\setlength{\tabcolsep}{4pt} 
\renewcommand{\arraystretch}{1.1} 
\begin{tabular}{lccc}
\hline
\textbf{Dataset} & \textbf{Total VMs\#} & \textbf{VMs\#} & \textbf{\%VMs} \\
\hline
\#1 AzurePublicDatasetV1 & $\sim$2.01M & $\sim$103k & 5.12\% \\
\#2 AzurePublicDatasetV2 & $\sim$2.70M & $\sim$180k & 6.65\% \\
\hline
\end{tabular}
\label{tab:2}
\end{table}

\subsection{Data Characteristics}

Typically, the collector records the minimum, average, and maximum workload consumption, along with corresponding timestamps (with a 5-minute time resolution). Please note that, for the V1 and V2 public datasets available in Table~\ref{tab:3}, vCPU historical workload data (resource measurement metrics) were unified to [0, 100] in Percentage [\%]. Thus, we considered average vCPU utilization for our Time-series analytics based on domain knowledge. Although historical Memory workload is unavailable in public datasets, VM memory (GBs) is available in the \texttt{VMtable}. Therefore, it can’t be used for time-series analytics, as it lacks memory footprint modeling, which enables multivariate forecasting and makes it more realistic for VM provisioning.

\begin{table}[h]
\centering
\caption{Measured VM metrics used as features within historical MS Azure VM workload data.}
\label{tab:vm_metrics}
\begin{tabular}{c|l|c}
\hline
\# & \textbf{Compute measurement metrics} & \textbf{Select} \\ \hline
1 & Timestamp [sec] & Yes \\
2 & Min CPU [\%] & No \\
3 & Max CPU [\%] & No \\
4 & Avg CPU [\%] & Yes \\ \hline
\end{tabular}
\label{tab:3}
\end{table}

%% file: Chapters/5_0_Methodology.tex
\section{METHODOLOGY}\label{sec:num5}
The method generates day-ahead probabilistic predictions for individual instance consumption time series. The approach consists of three sequential steps applied to each time series: Selection of VM profiles, Sanitation, and Conformal prediction. The method can be easily adapted to a specific dataset due to its modular approach. Based on the temporal budgets of public datasets, for large-scale forecasting, we allocate the first 75\% of each VM workload time series (approximately 3 weeks) to training and validation/calibration. The remaining 15\% (corresponding to the final week, typically 6–7 days depending on the workload lifetime) is reserved for testing, leveraging the longevity of the high-resolution data ($5$-minute sampling intervals). For medium-scale forecasting, the allocation ratio is set to 85\% for training/validation and 15\% for mid-term testing.

\begin{figure}[t]
\centering
\includegraphics[width=\columnwidth]{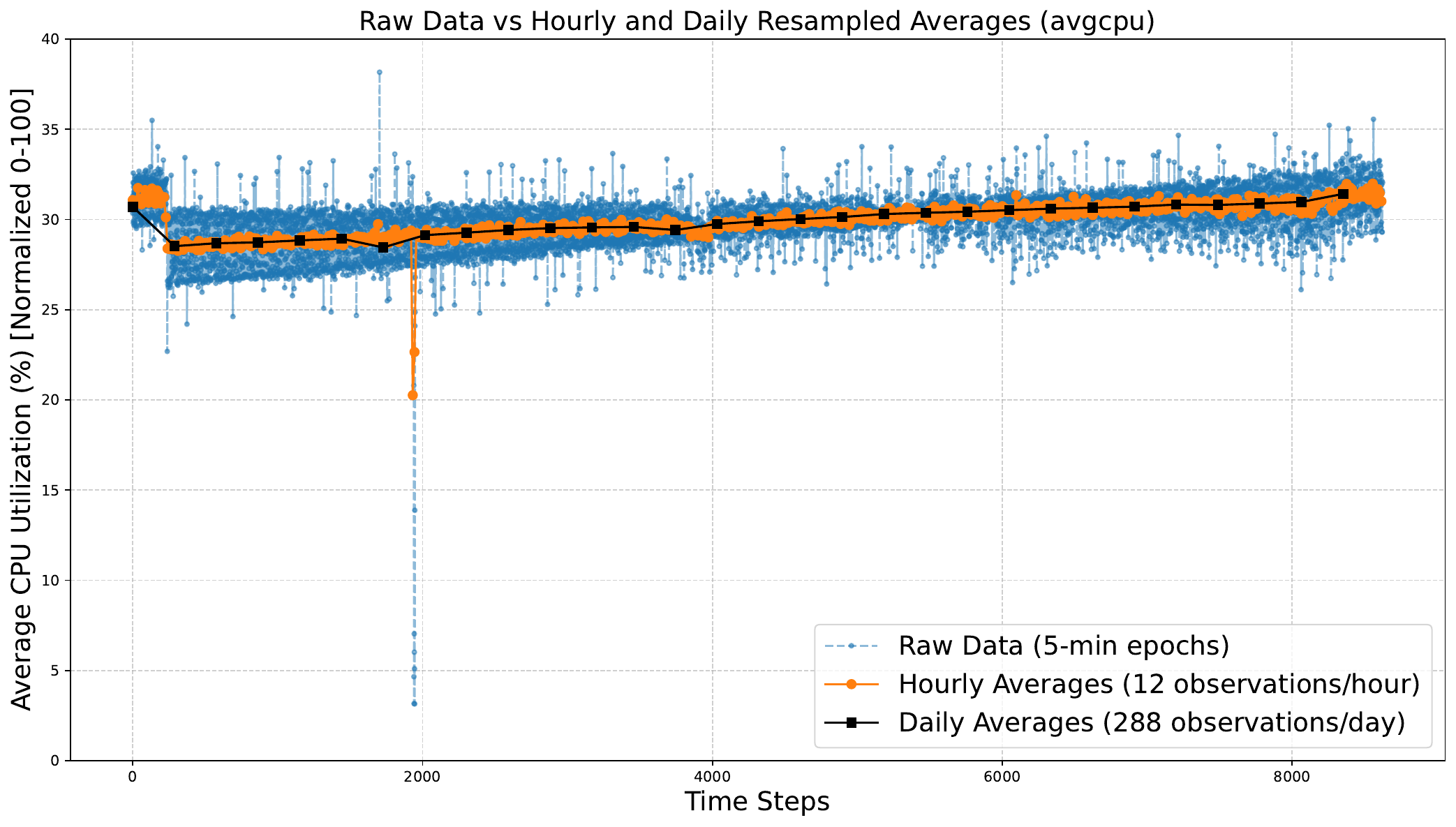}
\caption{Preserving 5-minute sampling intervals maintains the original workload variability, while hourly and daily resampling smooths patterns and risks losing critical VM profile information. VM workload utilization is commonly represented in a normalized form, scaled to $[0, 100]$ percent, relative to the virtual machine's provisioned compute or memory capacity.}
\label{fig:6}
\end{figure}


Figure~\ref {fig:6} illustrates the effect of temporal resampling on vCPU utilization patterns. The raw dataset, collected at 5-minute intervals, preserves fine-grained fluctuations that reflect the actual dynamics of the VM workloads. In contrast, hourly and daily resampling substantially smooth these patterns, attenuating variability and potentially obscuring short-term bursts in resource consumption. In our experiments, we avoided such resampling to prevent distortion of the underlying workload characteristics and to retain critical temporal information necessary for accurate VM profile analysis.

\subsection{VM Profile Selection and Characteristics}\label{sec:num4.2}

The following shows the VM category distributions before and after filtering for long-lifetime VMs that may be involved in our experiments. We categorized the VMs into three classes according to the core hour (\textit{corehours}) column within the data:  \textit{Delay-insensitive}, \textit{Interactive}, and \textit{Unknown}.

\begin{figure}[h]
\centering
\includegraphics[width=\columnwidth]{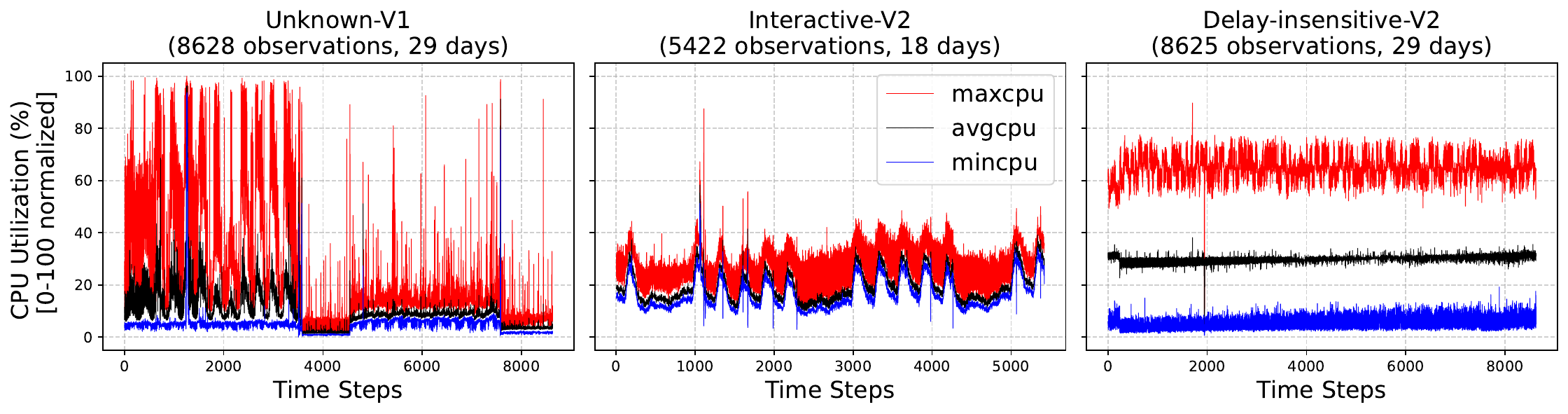}
\caption{Different classified virtual machine profiles with mid to long lifespans within Microsoft Azure's public datasets.}
\label{fig:8}
\end{figure}

\begin{figure}[h]
\centering
\includegraphics[width=\columnwidth]{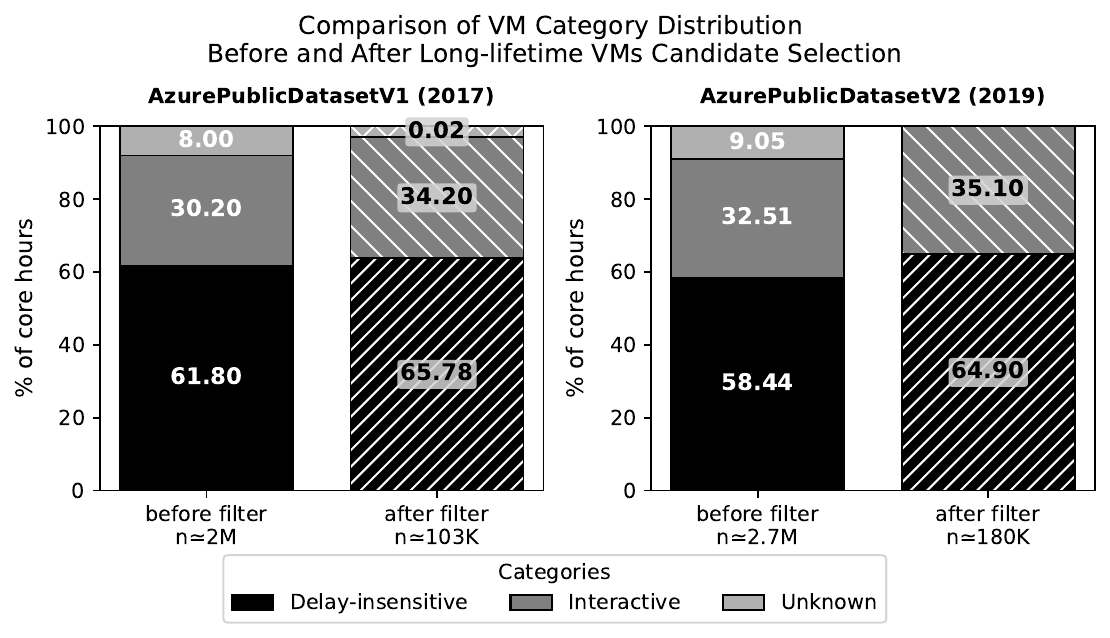}
\caption{Distribution of workload classes and their corresponding share of \textit{corehours} for the public Azure datasets V1 and V2, presented before and after long-lifetime VM candidates selection ($\geq$29 continuous days) and forecastability tests.}\label{fig:7}
\end{figure}

Figure~\ref{fig:7} shows that public datasets V1/V2 have more or less the same distribution classes, considering the percentage of core hours used. Also, after filtering long-lifetime VMs, there are a few \textit{Unknown} class VMs in \texttt{AzurePublicDatasetV2} (2019). All kinds of classified VM workload profiles are shown in Figure~\ref{fig:8} with different lifetimes during which VMs were created and terminated.

\subsection{Data Sanitation}\label{sec:num4.1}
To ensure data quality and mitigate noise sources reported in prior studies on virtual CPU (vCPU) telemetry, we implemented a systematic pre-processing pipeline before model training. First, short-term fluctuations in CPU usage measurements were de-noised using the Savitzky–Golay filter, which preserves local trends while smoothing high-frequency noise.\cite{bi2024arima, chen2022accurate} Second, a sequence integrity check was performed on the timestamp series to verify the expected 5-minute epoch intervals; any deviations indicated missing VM telemetry records, potentially arising from known data collection issues reported by the provider. \cite{cortez2019azureTelemetry} Third, these missing records (either single missing or a series of missing events, i.e., a gap) were imputed using the rolling-window median after exploring time-series decomposition test (trend, seasonality, noise) and different imputation solutions (various interpolation techniques and modern way such as Self-attention-based imputation for time series (SAITS) proposed by \cite{du2023saits}) in context of the missing data problem in time-series, thereby preserving temporal continuity while avoiding bias from extreme values as well as cloud failure prediction.\cite{yang2023diffusion} Finally, we identified and corrected overshoot consumption events and other anomalous spikes by applying Z-score-based outlier detection followed by value clipping. This procedure reduces the influence of transient measurement artifacts that could distract the learning process and degrade generalization performance in the forecasting models. Our pre-processing experiments are available on our GitHub repository\footnote{\scriptsize \url{https://github.com/clevilll/MS-Azure-VM-WL-CP-Forecast-Characterization/}}

%% file: Chapters/5_1_Problem_Statement.tex
\subsection{Problem Statement}
Hyperscalers' environments pose unique challenges for right-sizing VMs due to the dynamic, unpredictable nature of cloud workloads. Classic right-sizing approaches often rely on static or simplistic models that do not fully account for the variability in VM utilization over time. This problem addresses the need for predictive analytics techniques that can accurately forecast mid- to long-term workload patterns and provide actionable RSR. The objective is to develop intelligent algorithms that can predict future utilization trends, identify correlations across multiple VMs, and potentially recommend VM sizes that optimize both performance and cost-efficiency for hyperscaler users.

In this research, we aim to address VM workload characterization for the Right‑Sizing Recommendation (RSR) use case in the context of large‑scale workload forecasting by exploring the potential of time-series-friendly learning ML models (leveraging backtesting for model evaluation) to learn VM workload behavior and empower them by PI that returns the predicted lower/upper-bound of a time-series in a specific time horizon without relying on complex transformations (e.g., decomposition, FFT). The goal is to reduce over-provisioning and under-provisioning through conformal predictive ML models, thereby improving the overall efficiency of cloud operations. In the current study, we focused solely on vCPU usage and its forecasting for our investigation, as this information is readily available in public datasets. Therefore, we are working on PI and UB prediction using historical workload data. In other words, our analysis is limited to univariate time-series analytics (U-TSA); thus, we seek smart Univariate Forecasting (UF) models tailored to the workload for the RSR use case. Given univariate time data, we consider only one workload variable that varies over time. It exhibits non-stationary behavior, including potential periodic fluctuations/patterns as well as seasonality at regular intervals due to maintenance-related factors or weekends. Irregular behavior can also be caused by security-related factors, such as Cryptojacking \cite{nissar2024novel, jayasinghe2020survey}, participation in Botnets/DDoS attacks \cite{agrawal2019defense,somani2017ddos}, and the Misuse of idle VMs.

The classic UF is insufficient to comprehend, model, and predict the behavior of VM workload variables over time, especially without Backtesting (BT) to update learning or retrain models. Also, it is not possible to learn complex workload behavior using high-frequency time data. Thus, we need models that are capable of making skillful forecasts compared to naive models and tuned SARIMA models on univariate time series forecasting problems that have both trend and seasonal components, without using cross-validation techniques meaningfully \cite{bergmeir2018note} like BT, as this could be expensive for RSR to apply on millions of instances. 

%% file: Chapters/8_discussion_Results.tex
\section{Discussion and Results}\label{sec:num8}

\subsection{Global top-tier models for long-lived VM workload profiles}
After identifying the global top-tier models for long-lived VM workload profiles, we ranked the results (bar charts) for the top 1 and top 3 models:
\begin{figure}[h]
\centering
\includegraphics[width=\columnwidth]{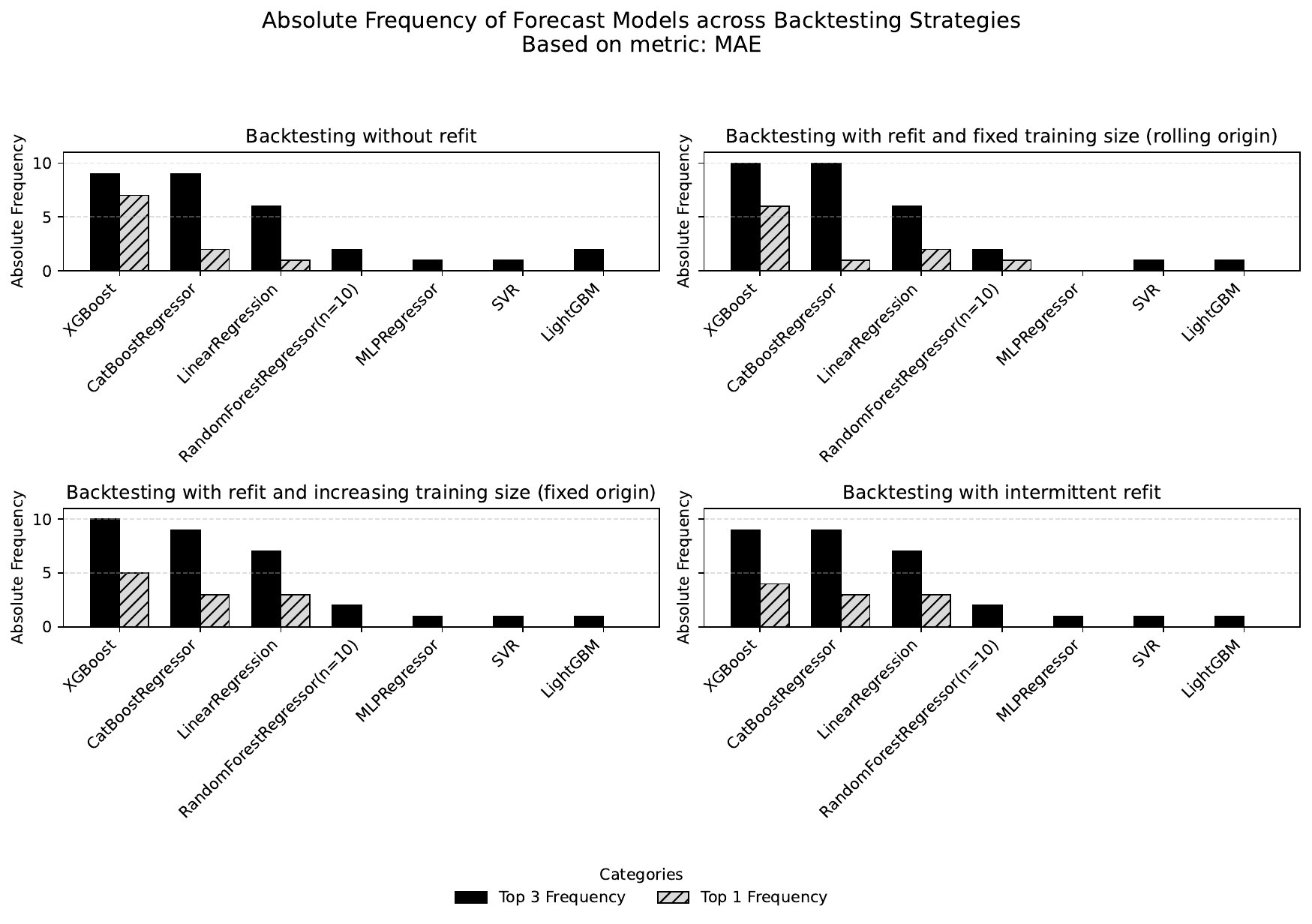}
\caption{We ranked ML-powered conformal forecasts based on the upper bound (UB) within PI over limited VM candidates evaluated by the MAE metric using Backtesting with refit for vCPU for the top 1 and top 3 models categories among limited VM candidates due to expensive computation, specifically for large forecast scenarios for some ML models concerning the used BT-based approach (re-training) and estimate PI using bootstrapping on high-resolution time data.}
\label{fig:11}
\end{figure}
Across all backtesting strategies, \texttt{XGBoost} and \texttt{CatBoostRegressor} consistently dominate, achieving the highest Top 3 and frequent Top 1 rankings, demonstrating robust accuracy across varying temporal splits.  \texttt{LinearRegression} performs moderately well in some setups but rarely leads, while \texttt{RandomForestRegressor}, \texttt{MLPRegressor}, \texttt{LightGBM}, and \texttt{SVR} appear less frequently and show limited top performance. Refitting strategies slightly improve Top 1 placements, suggesting that incorporating recent data enhances short-term forecast accuracy, though the overall performance hierarchy remains unchanged.

Likely, the results of ranked ML-powered conformal forecasts based on the Prediction Interval (PI), evaluated by the Prediction Interval Coverage Probability (PICP) metric, again show the dominance of \texttt{XGBoost} and \texttt{CatBoostRegressor} across all backtesting strategies. However, ranked ML-powered conformal forecasts based on the target values evaluated by the MAE metric show that \texttt{LinearRegression} is the top model, regardless of using the backtesting strategy.

\subsection{Efficiency models for long-lived VM workload profiles}
\begin{figure}[h]
\centering
\includegraphics[width=\columnwidth]{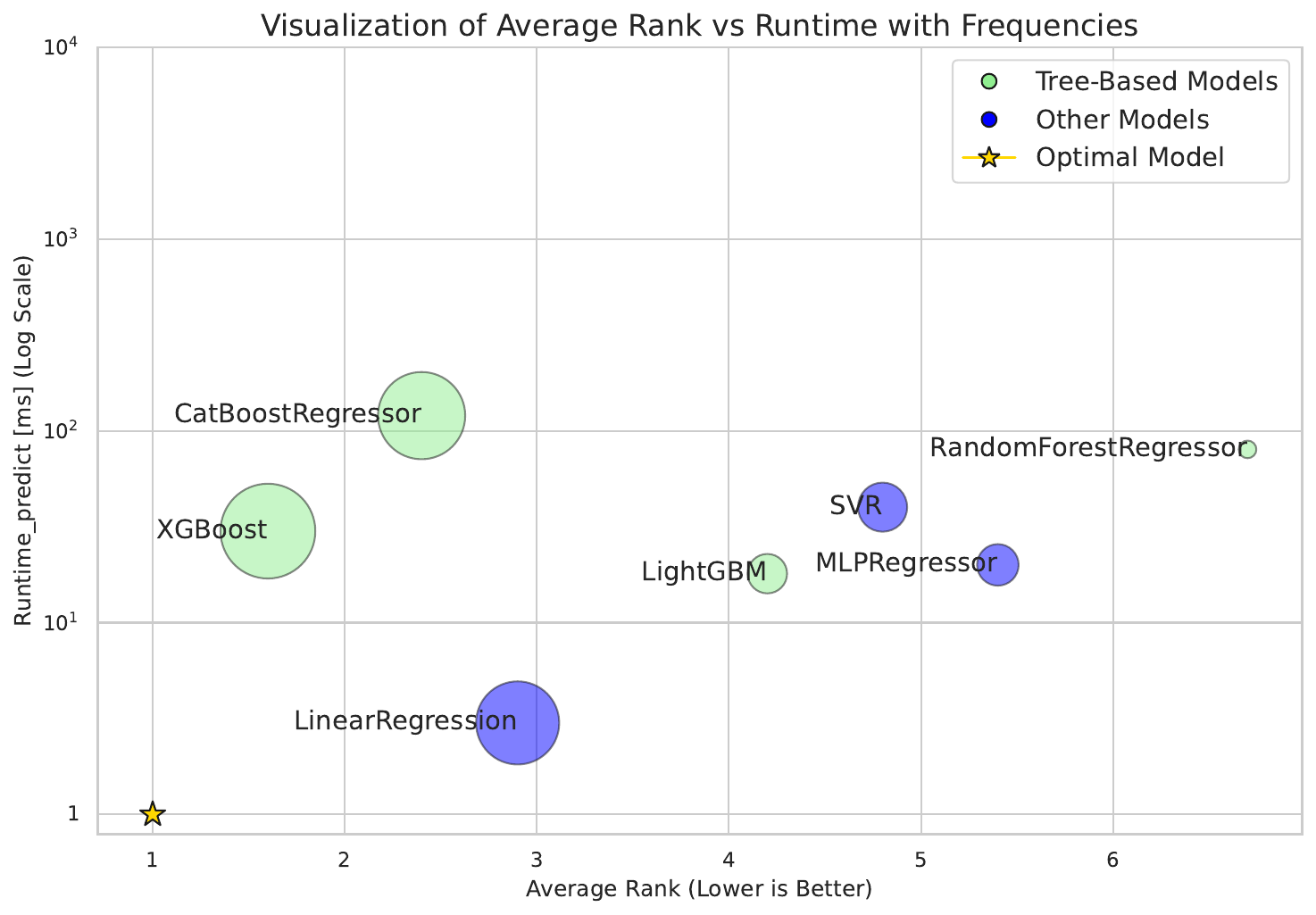}
\caption{We ranked VM candidates using Backtesting with refit and fixed training size for upper bound (UB) estimation on CPU/GPU. \texttt{CatBoostRegressor} and \texttt{XGBoost} were the most efficient, while \texttt{RandomForestRegressor} had high run-time, which CUDA and CuML could optimize on GPUs. Further ranking results (bar charts) for the top 1 and top 3 models are available in our GitHub%
    \ifdef{\href}{}{.}%
    \ifdefined\href  \href{https://github.com/clevilll/MS-Azure-VM-WL-CP-Forecast-Characterization/blob/main/README.md}{ repository}\fi.}\label{fig:10}
\end{figure}

After identifying the global top-tier models for long-lived VM workload profiles, the plot in Figure~\ref{fig:10} compares predictive accuracy (average rank) against runtime for UB estimation. \texttt{CatBoostRegressor} and \texttt{XGBoost} achieve the best balance of high accuracy and low runtime, making them efficient choices for large-scale VM forecasting. \texttt{RandomForestRegressor} performs well in accuracy but incurs high runtime, while \texttt{LinearRegression} is the fastest but less accurate. Other models fall between these extremes, showing trade-offs between speed and accuracy.
\subsection{Limitations}
Despite the breadth of analysis conducted in this study and anonymized timestamps in public datasets that avoid design features using \textit{Time-based} features (i.e, calendar information, elapsed time, Fourier terms, etc.) that respect temporal structure, several limitations remain.  
First, the inherent unpredictability of time series data in large-scale cloud environments poses challenges for robust forecasting. \cite{zadeh2025towards} Such variability, driven by dynamic and often irregular workload patterns, should be explicitly quantified to better assess model reliability under uncertainty.  Second, the absence of critical service-level metadata (e.g., the types, qualities, and quantities of hosted web applications) limits deeper insights into VM workload behavior. This lack of contextual information limits the ability to build richer VM workload profiles. It constrains achievable forecasting accuracy because the models cannot fully account for application-specific performance drivers.

%% file: Chapters/9_future-work-and-conclusion.tex
\section{Conclusion \& Future Work}\label{sec:num9}
Operating cloud infrastructure efficiently, especially in large cloud service provider (CSP) or hyperscaler environments, requires optimizing physical resource utilization to minimize costs and maximize performance. Efficient cloud infrastructure operation requires optimal VM sizing to avoid costly over- or under-provisioning, especially for large-scale cloud applications \cite{rebjock2020simple}. In this work, we applied minimal-impact preprocessing, including Savitzky–Golay filtering, to preserve the natural characteristics of high-frequency VM workload data while reducing noise. Using conformal prediction (CP), we generated reliable prediction intervals for medium- and long-term forecasts, enabling innovative and efficient provisioning decisions, with further validation based on available VM sizes and computational capacities. Ranked results, including cross-validation using BT strategies across multiple VM profiles for the UB scenario, demonstrate that advanced tree-based algorithms perform well for long-horizon forecasting, although other models achieve comparable accuracy. However, the absence of critical service-level metadata limits deeper workload behavior analytics, profiling, and potentially reduces forecasting accuracy. Future work will combine Temporal Conformal Prediction (TCP) \cite{aich2025temporal} for adaptive intervals on non-stationary workloads with trend forecasting via decomposition methods (e.g., Wavelet Reconstruction) or other existed visual representations to capture long-term vCPU patterns, including memory footprint to handle today’s non-stationary, regime-shifting, and rapidly changing VM utilization patterns by wrapping it with \textit{Resource Scaling Rules} and realistic intelligent VM provisioning in hyperscaler environments.